\documentclass[final]{cvpr}

\usepackage{times}
\usepackage{epsfig}
\usepackage{graphicx}
\usepackage{amsmath}
\usepackage{amssymb}
\usepackage{multirow}
\usepackage{algorithm}
\usepackage{algorithmic}
\usepackage[pagebackref=true,breaklinks=true,colorlinks,bookmarks=false]{hyperref}
\usepackage{subfigure}

\def\eg{\emph{e.g.}}
\def\ie{\emph{i.e.}}
\def\etc{\emph{etc}}

\newif\ifsubmit
\submitfalse
\ifsubmit
\newcommand{\aishan}[1]{}
\else
\newcommand{\aishan}[1]{\textcolor{blue}{[aishan: #1]}}
\fi

\def\cvprPaperID{1610} 
\def\confYear{CVPR 2021}

\begin{document}

\title{Dual Attention Suppression Attack: Generate Adversarial \\Camouflage in Physical World}

\author{Jiakai Wang , Aishan Liu, Zixin Yin, Shunchang Liu ,\\
Shiyu Tang, and Xianglong Liu\textsuperscript{\thanks{Corresponding author}}\\
\fontsize{11.0pt}{\baselineskip}\selectfont State Key Lab of Software Development Environment, \\Beihang University, Beijing, China\\
{\tt\small \{jk\_buaa\_scse, liuaishan, yzx835, liusc, sytang, xlliu\}@buaa.edu.cn}}
\maketitle
\begin{abstract}Deep learning models are vulnerable to adversarial examples. As a more threatening type for practical deep learning systems, physical adversarial examples have received extensive research attention in recent years. However, without exploiting the intrinsic characteristics such as model-agnostic and human-specific patterns, existing works generate weak adversarial perturbations in the physical world, which fall short of attacking across different models and show visually suspicious appearance.
Motivated by the viewpoint that attention reflects the intrinsic characteristics of the recognition process, this paper proposes the Dual Attention Suppression (DAS) attack to generate visually-natural physical adversarial camouflages with strong transferability by suppressing both model and human attention. As for attacking, we generate transferable adversarial camouflages by distracting the model-shared similar attention patterns from the target to non-target regions. Meanwhile, based on the fact that human visual attention always focuses on salient items (e.g., suspicious distortions), we evade the human-specific bottom-up attention to generate visually-natural camouflages which are correlated to the scenario context. We conduct extensive experiments in both the digital and physical world for classification and detection tasks on up to date models (e.g., Yolo-V5) and significantly demonstrate that our method outperforms state-of-the-art methods.\footnote{Our code can be found in \url{https://github.com/nlsde-safety-team/DualAttentionAttack}.}
\end{abstract}

\section{Introduction}
Deep neural networks (DNNs) have achieved remarkable performance across a wide areas of applications, \eg, computer vision \cite{Krizhevsky2012ImageNet}, natural language \cite{sutskever2014sequence}, and acoustics \cite{speech}, \etc, but they are vulnerable to \emph{adversarial examples} \cite{szegedy2013intriguing}. These elaborately designed perturbations are imperceptible to humans but can easily lead DNNs to wrong predictions, which pose a strong security challenge to deep learning applications in both the digital and physical world \cite{goodfellow,Eykholt_2018_CVPR,uapaco}.

\begin{figure}[!htb]
    \centering
\subfigure[]{
\includegraphics[width=0.45\linewidth]{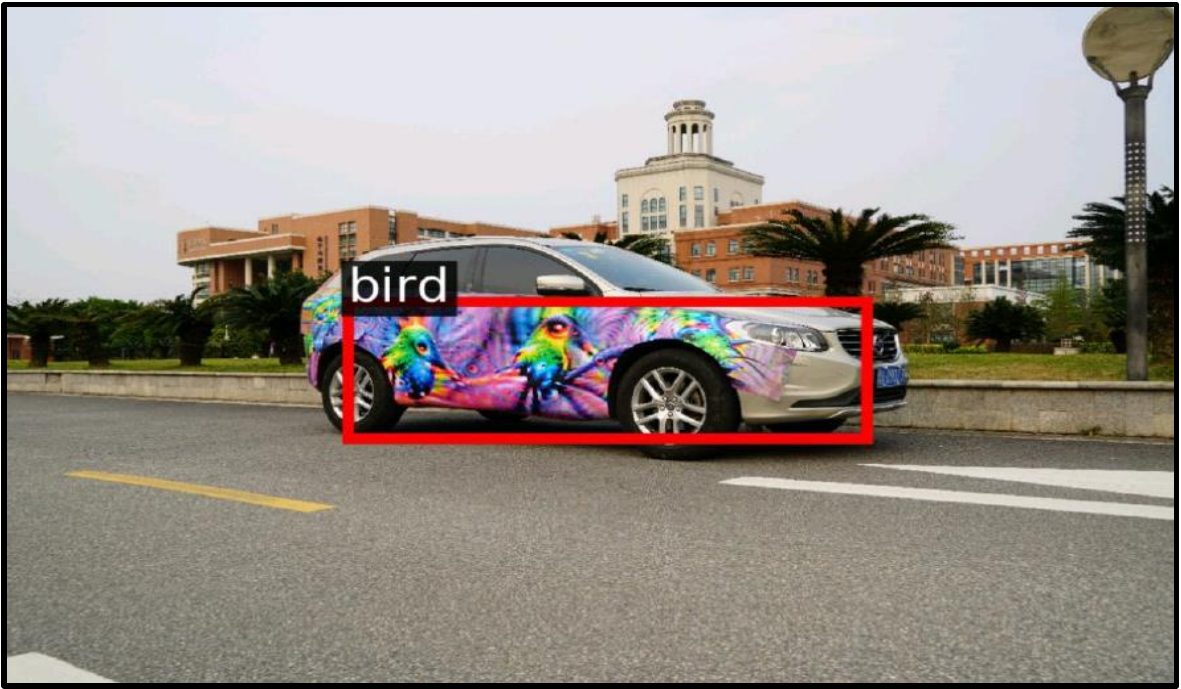}
  \label{fig:rst1}
}
\subfigure[]{
\includegraphics[width=0.45\linewidth]{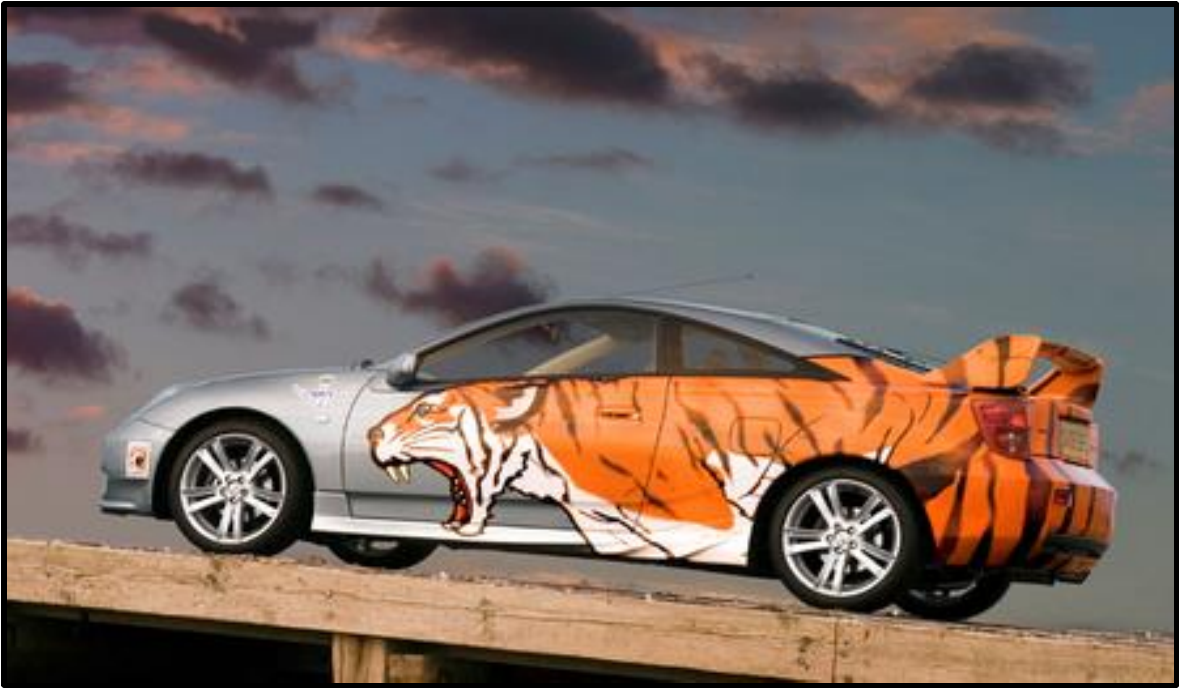}
  \label{fig:rst2}
}
\subfigure[]{
\includegraphics[width=0.45\linewidth]{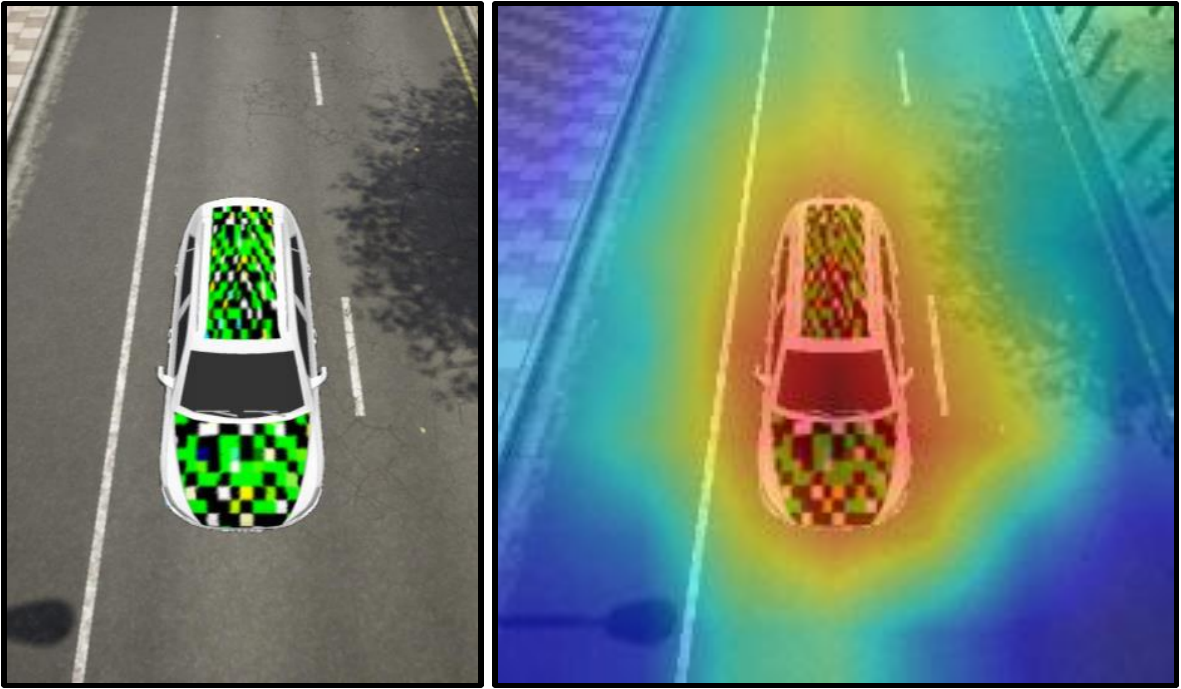}
  \label{fig:rst3}
}
\subfigure[]{
\includegraphics[width=0.45\linewidth]{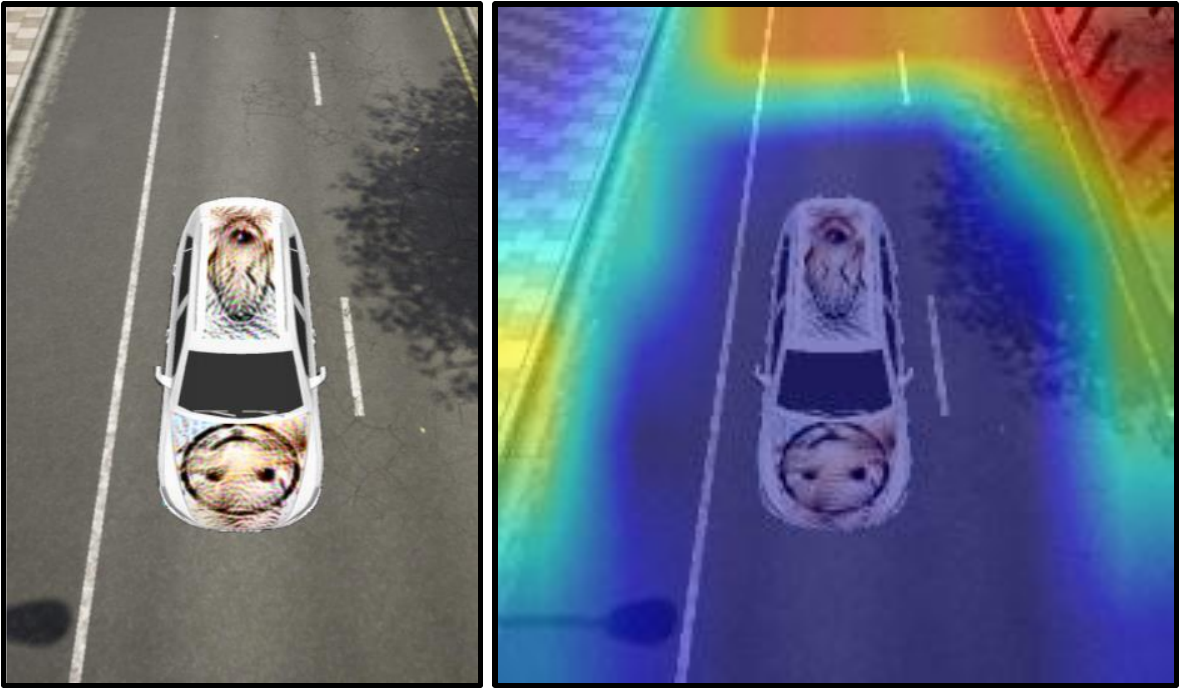}
  \label{fig:rst4}
}
\caption{(a) shows the suspicious appearance of camouflages generated by previous work (\ie, {UPC} \cite{Huang_2020_CVPR}). (b) is the painted car that commonly exists in the physical world. (c) shows the adversarial example (classified as \texttt{pop bottle}) generated by existing work (\ie, CAMOU \cite{zhang2018camou}) and its corresponding attention map. (d) shows the adversarial example (classified as \texttt{Shih-Tzu}) generated by our DAS and its distracted attention map.}
\label{fig:app-intro}
\end{figure}

In the past years, a long line of work has been proposed to perform adversarial attacks in different scenarios under different settings \cite{kurakin2016adversarial,dong2017boosting,pmlr-v80-athalye18b}. Though challenging deep learning, adversarial examples are also valuable for understanding the behaviors of DNNs, which could provide insights into the blind-spots and help to build robust models \cite{ilyas2019adversarial,tsipras2018robustness,Li2021Understanding,zhang2020interpreting}. Generally, adversarial attacks can be divided into two categories: \emph{digital attacks}, which attack DNNs by perturbing the input data in the digital space; and \emph{physical attacks}, which attack DNNs by modifying the visual characteristics of the real object in the physical world. In contrast to the attacks in the digital world \cite{yunhan_2019_CoRR, Xie_2019_CVPR, Inkawhich_2019_CVPR, zhang2018camou}, adversarial attacks in the physical world are more challenging due to the complex physical constraints and conditions (\eg, lighting, distance, camera, \etc.), which will impair the attacking ability of generated adversarial perturbations \cite{NIPS2018_7647}. In this paper, we mainly focus on the more challenging physical world attack task, which is also more meaningful to the deployed deep learning applications in practice.

Though several attempts have been adopted to perform physical attacks \cite{uapaco, Huang_2020_CVPR, liu2019perceptual}, existing works always ignore the intrinsic characteristics such as model-agnostic and human-specific patterns so that their attacking abilities are still far from satisfactory. In particular, the limitations can be summarized as (1) the existing methods ignore the common patterns among models and generate adversarial perturbations using model-specific clues (\eg, gradients and weights of a specific model), which fails to attack across different target models. In other words, the transferability of adversarial perturbations is weak, which impairs their attacking abilities in the physical world; (2) current methods generate adversarial perturbations with a visual suspicious appearance which is poorly aligned with human perception and even attracts the human attention. 
For example, painted on the adversarial camouflage \cite{Huang_2020_CVPR}, the classifier misclassifies the car into a bird. However, as shown in Figure \ref{fig:rst1}, the camouflage apparently contains un-natural and suspicious bird-related features (\eg, bird head), which attracts human attention.

To address the mentioned problems, this paper proposes the Dual Attention Suppression (DAS) attack by suppressing both the model and human attention.
Regarding the \textbf{transferability for attacks}, inspired by the biological observation that cerebral activities between different individuals share similar patterns when stimulus features are encountered \cite{Zatorre_attention} (\ie, selected attention \cite{Tricoche_peer}), we perform adversarial attacks by suppressing the attention patterns shared among different models. Specifically, we distract the model-shared similar attention from target to non-target regions via connected graphs. Thus, target models will be misclassified by not paying attention to the objects in the target region. Since our generated adversarial camouflage captures model-agnostic structures, it can transfer among different models, which improves the transferability.

As for the \textbf{visual naturalness}, psychologists have found that the bottom-up attention of human vision will alert people to salient objects (\eg, distortion) \cite{CONNOR2004R850}. Existing methods generate physical adversarial examples with visually suspicious appearance, which shows salient features to human perception. Thus, we try to evade this human-specific visual attention by generating adversarial camouflage which contains high semantic correlation to scenario context. As a result, the generated camouflage is more unsuspicious and natural in terms of human perception. Figure \ref{fig:rst3} is the adversarial camouflage generated by CAMOU \cite{zhang2018camou} which is suspicious to human vision. By contrast, our generated adversarial camouflage yields a more natural appearance as shown in Figure \ref{fig:rst4}.

To the best of our knowledge, we are the first to exploit the shared attention characteristics among models and generate adversarial camouflage in the physical world by suppressing both the model and human attention. Extensive experiments in both the digital and physical world on both classification and detection tasks are conducted which demonstrate that our method outperforms other state-of-the-art methods.

\section{Related Works}
Adversarial examples are elaborately designed perturbations which are imperceptible to human but could mislead DNNs \cite{szegedy2013intriguing,goodfellow}. In the past years, a long line of work has been proposed to develop adversarial attack strategies \cite{KurakinGB16,Eykholt_2018_CVPR,liu2019perceptual,autocheckout,Duan_2020_CVPR,liu_2020_eccv,zhang2018camou,Huang_2020_CVPR}. In general, there are several different ways to categorize adversarial attack methods, \eg, targeted or untargeted attacks, white-box or black-box attacks, \etc. Based on the domain in which the adversarial perturbations are generated, adversarial attacks can be divided into digital attacks and physical attacks. 

Digital attacks generate adversarial perturbations for input data in the digital pixel domain. Szegedy \etal. \cite{szegedy2013intriguing} first introduced adversarial examples and used the L-BFGS method to generate them. By leveraging the gradients of the target model, Goodfellow \etal. proposed the Fast Gradient Sign Method (FGSM) \cite{goodfellow} which could generate adversarial examples quickly. Moreover, Madry \etal. \cite{madry_towards} proposed Projected Gradient Decent (PGD), which is currently the strongest first-order attack method. Based on the gradient of the target model, a series of attack approaches have been proposed \cite{KurakinGB16,Dong_2018_CVPR,Xie_2019_CVPR,Dong_2019_CVPR}. Although these attacks achieve substantial results in the digital world, their attacking abilities degenerate significantly when introduced into the physical world.

On the other hand, physical attacks aim to generate adversarial perturbations by modifying the visual characteristics of the real object in the physical world. To achieve the goal, several works first generate adversarial perturbations in the digital world, then perform physical attacks by painting the adversarial camouflage on the real object or directly create the perturbed objects. By constructing a rendering function, Athalye \etal. \cite{pmlr-v80-athalye18b} generated 3D adversarial objects in the physical world to attack classifiers. Eykholt \etal. \cite{Eykholt_2018_CVPR} introduced NPS \cite{nps} into the loss function which considers the fabrication error so that they can generate strong adversarial attacks for traffic sign recognition. Recently, Huang \etal. \cite{Huang_2020_CVPR} proposed the Universal Physical Camouflage Attack (UPC), which crafts camouflage by jointly fooling the region proposal network and the classifier. 
Another line of work tries to perform physical adversarial attacks by generating adversarial patches \cite{advpatch}, which confine the noise to a small and localized patch without perturbation constraint \cite{liu2019perceptual, uapaco}.

\section{Approach}
In this section, we first provide the definition of the problem and then elaborate on our proposed framework.

\subsection{Problem Definitions}

Given a deep neural network $\mathbb{F_{\theta}}$ and an input clean image $\mathbf{I}$ with the ground truth label $y$, an adversarial example $\mathbf{I}_{adv}$ in the \textbf{digital world} can make the model conduct wrong predictions as follows:
 \begin{equation}
    \begin{split}
        \mathbb{F}_{\theta}(\mathbf{I}_{adv}) \neq y \quad s.t.\quad  \lVert\mathbf{I} - \mathbf{I}_{adv}\rVert < \epsilon
        ,
    \end{split}
\end{equation}
where $||\cdot||$ is a distance metric to quantify the distance between the two inputs $\mathbf{I}$ and $\mathbf{I}_{adv}$ sufficiently small.

In the \textbf{physical world}, let $(\mathbf{M}, \mathbf{T})$ denote a 3D real object with a mesh tensor $\mathbf{M}$, a texture tensor $\mathbf{T}$, and ground truth $y$. The input image $\mathbf{I}$ for a deep learning system is the rendered result of the real object $(\mathbf{M}, \mathbf{T})$ with environmental condition $c \in \mathbf{C}$ (\eg, camera views, distance, illumination, \etc.) from a renderer $\mathcal{R}$ by $\mathbf{I} = \mathcal{R}((\mathbf{M}, \mathbf{T}), c)$. To perform physical attacks, we generate $\mathbf{I}_{adv} = \mathcal{R}((\mathbf{M}, \mathbf{T}_{adv}), c)$ through replacing the original $\mathbf{T}$ with an adversarial texture tensor $\mathbf{T}_{adv}$, which has different physical properties (\eg, color, shape).
Thus the definition of our problem can be depicted as:
\begin{equation}
    \begin{split}
        \mathbb{F}_{\theta}(\mathbf{I}_{adv}) \neq y \quad \emph{s.t.}\quad\lVert \mathbf{T} - \mathbf{T}_{adv} \rVert < \epsilon,
    \end{split}
\end{equation}
where we ensure the naturalness of the generated adversarial camouflage in the physical world by $\epsilon$.

In this paper, we mainly discuss adversarial attacks in the physical world and generate an adversarial camouflage (\ie, texture), which is able to fool the real deep learning systems when it is painted or overlaid on a real object.

\subsection{Framework Overview}
\begin{figure*}[!htb]
\begin{center}
\includegraphics[width=0.9\linewidth]{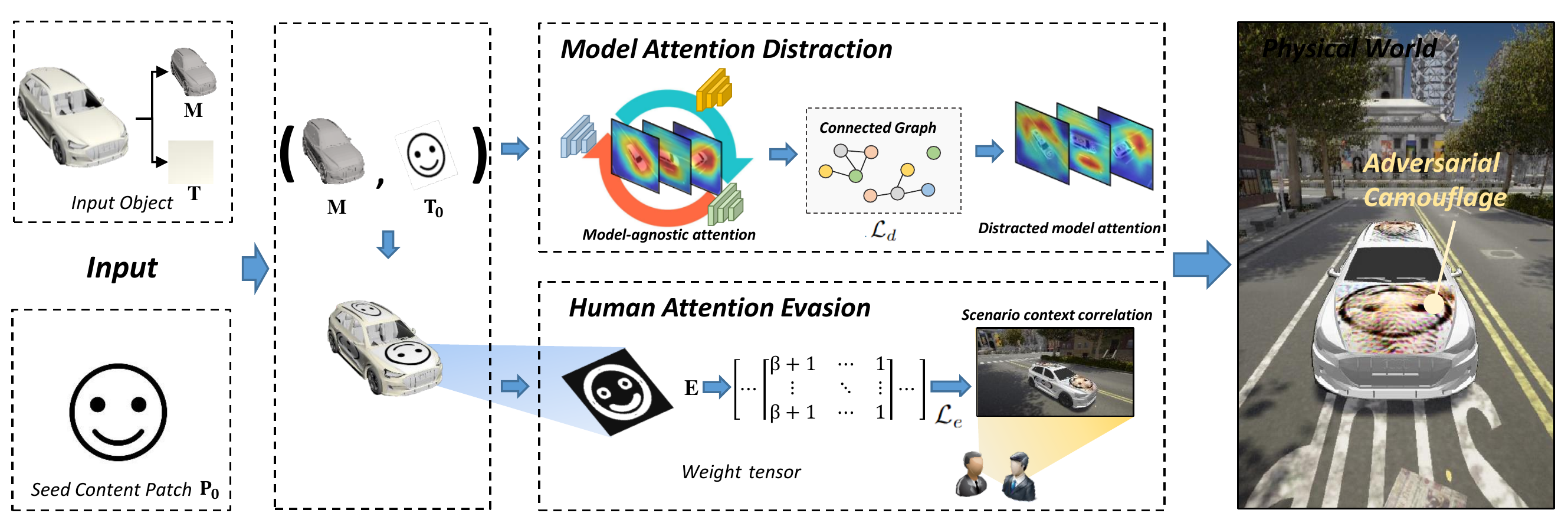}
\end{center}
 \caption{The framework of our DAS method. We first distract the intrinsic attention characteristic through fully exploiting the similar attention patterns of models and forcing the ``heat" regions away from the target object with loss function $\mathcal{L}_{d}$. Then we evade the human-specific visual attention mechanism by correlating the appearance of adversaries to the context scenario and preserving the shape information of seed content image to generate visually-natural adversarial camouflage.}
\label{fig:framework}
\end{figure*}

To generate visually-natural physical adversarial camouflage with strong transferability, we propose the Dual Attention Suppression (DAS) framework which suppresses both the model and human attention. The overall framework can be found in Figure \ref{fig:framework}.

Regarding the \textbf{transferability for attack}, inspired by the biological observation, we suppress the similar attention patterns shared among models. Specifically, we generate adversarial camouflage by distracting the model attention from target to non-target regions (\eg, background) via connected graphs. Since different deep models yield similar attention patterns towards the same object, our generated adversarial camouflage could capture the model-agnostic structures and transfer to different models.

As for the \textbf{visual naturalness}, we aim to evade the human-specific bottom-up attention in human vision \cite{CONNOR2004R850} by generating visually-natural camouflage. By introducing a seed content patch $\mathbf{P}_{0}$, which has a strong perceptual correlation to the scenario context, the generated adversarial camouflage in this case can be more unsuspicious and natural to human perception. Since humans pay more attention to object shapes when making predictions \cite{liu_2020_eccv}, we further preserve the shape information of the seed content patch to improve the human attention correlations. Thus, the human-specific attention mechanism is evaded, leading to more natural camouflage.

\subsection{Model Attention Distraction}

Biologists have found that the same stimulus features (\ie, selected attention) yield similar patterns of cerebral activities among different individuals \cite{Zatorre_attention} (\ie, similar characteristics of the neuron hyper-perception). Since artificial neural networks are implemented from the human central nervous system \cite{Michael_cure}, it is also reasonable for us to assume that DNNs may have the same characteristics, \ie, different models have similar attention patterns towards the same objects when making the same predictions. Based on the above observations, we consider improving the transferability of adversarial camouflages by capturing the model-agnostic attention structures.

Visual attention techniques \cite{Zhou_2016_CVPR} have been long studied to improve the explanation and understanding of deep learning behaviors, such as CAM \cite{Zhou_2016_CVPR}, Grad-CAM \cite{Selvaraju_2017_ICCV}, and Grad-CAM++ \cite{gradcampp}. When making predictions, a model pays most of its attention to the target objects rather than meaningless parts. Intuitively, to successfully attack a model, we directly distract the model attention from the salient objects. In other words, we distract the model-shared similar attention map on the salient area to other regions and force the attention weights to distribute uniformly through the entire image. Thus, the model may fail to focus on the target object and make the wrong predictions.

Specifically, given an object $(\mathbf{M}, \mathbf{T})$, an adversarial texture tensor $\mathbf{T}_{adv}$ to be optimized, and a certain label $y$, we get $\mathbf{I}_{adv}$ by $\mathcal{R}$ and then compute the attention map $\mathbf{S}^{y}$ with an attention module $\mathcal{A}$ as

\begin{equation}
\begin{split}
    \mathbf{S}^{y} = \mathcal{A}(\mathbf{I}_{adv}, y).\\
\end{split}
\label{eqn:atten}
\end{equation}

More precisely, the attention module $\mathcal{A}$ is
\begin{equation}
    \begin{split}
        \mathcal{A}(\mathbf{I},y) = \emph{relu}(\sum_{k}\sum_{i}\sum_{j}\alpha^{ky}_{ij}\cdot\emph{relu}(\frac{\partial p^y}{\partial A^{k}_{ij}})\cdot A^{k}_{ij}),
    \end{split}
\end{equation}
where $\alpha^{ky}_{ij}$ is the gradient weights for a particular class $y$ and activation map $k$, $p^y$ is the score of the class $y$, $A^{k}_{ij}$ is the pixel value in position $(i,j)$ of the $k$-th feature map, and $\emph{relu}(\cdot)$ denotes the $\emph{relu}$ function. Note that the attention module can be an arbitrary deep learning model rather than the target model.

Given the attention map $\mathbf{S}^y$ calculated by Eqn \ref{eqn:atten}, we aim to distract the attention region and force the model to focus on non-target regions. Intuitively, the pixel value of the attention map represents to what extent the region contributes to model predictions. To decrease the attention weights of the salient object and disperse these attention regions, we exploit the \emph{connected graph}, which contains a path between any pair of nodes within the graph. In an image, a region with attention weights for each pixel higher than a specific threshold can be deemed as a connected region. To distract the model attention using the connected graph, we consider the following two tasks: (1) decrease the overall connectivity by separating connected graphs into multiple sub-graphs; (2) reduce the weight of each node within a connected sub-graph. To achieve these goals, we propose attention distraction loss as
\begin{equation}
    \begin{split}
        \mathcal{L}_{d} =
        \frac{1}{K}\sum_{k}\frac{G_{k}}{N-N_k}, \quad
        \emph{s.t.} \quad G_{k} \subseteq \mathbf{S}^y,
    \end{split}
    \label{eqn:adloss}
\end{equation}
where $G_{k}$ is the sum of pixel values in the region corresponding to $k$-th connected graph in $\mathbf{S}^y$, $N$ is the total pixel number of the $\mathbf{S}^y$, and $N_k$ is the total pixel number of $G_{k}$. By minimizing $\mathcal{L}_{d}$, the salient region in the attention map becomes smaller (\ie, distracted) and the pixel values of the salient regions become lower (\ie, no longer ``heated''), leading to the ``distracted'' attention map.

\subsection{Human Attention Evasion}
\label{sec:content}

To overcome the problem brought by the complex environmental conditions in the physical world, most physical attacks generate adversarial perturbations with a comparatively huge magnitude \cite{Duan_2020_CVPR}. Since the bottom-up human attention mechanism always alerts people to salient objects (\eg, distortion) \cite{CONNOR2004R850}, adversarial examples in this case can always attract human attention due to the salient perturbations, showing suspicious appearance and lower stealthiness in the physical world.  

In this paper, we aim to generate more visually-natural camouflage by suppressing the human visual mechanism, which will evade human-specific attention. 
Intuitively, we expect the generated camouflage to share similar visual semantics with the context to be attacked (\eg, beautiful paintings on vehicles are more perceptually acceptable for humans than meaningless distortions). Thus, the generated adversarial camouflage can be highly correlated to human perception, which is unsuspicious to human perception. 

In particular, we first incorporate a seed content patch $\mathbf{P}_{0}$ which contains a strong semantic association with the scenario context. We then paint the seed content patch on the vehicle $(\mathbf{M},\mathbf{T})$ by $\mathbf{T}_{0} = \Psi(\mathbf{P}_{0},\mathbf{T})$. Specifically, $\Psi(\cdot)$ is a transformation operator which first transfers the 2D seed content patch into a 3D tensor, and then paint the car through tensor addition.

Since humans pay more attention to shapes when focusing on objects and making predictions \cite{liu_2020_eccv}, we aim to further improve the human attention correlation by better preserving the shape of the seed content patch. Specifically, we obtain the edge patch $\mathbf{P}_{edge} = \mathbf{\Phi}(\mathbf{P}_{0})$ using an edge extractor $\mathbf{\Phi}$ \cite{Canny4767851} from the seed content patch. It should be noticed that $\mathbf{P}_{edge}$ has 0-1 value in each pixel. After that, we simply transform the edge patch $\mathbf{P}_{edge}$ to a mask tensor $\mathbf{E}$ which has the same dimension with $\mathbf{T}_0$.

With mask tensor $\mathbf{E}$, we can distinguish the edge and non-edge regions and limit the perturbations added to the edge regions. Thus, the attention evasion loss $\mathcal{L}_{e}$ can be formulated as

\begin{equation}
    \begin{split}
        \mathcal{L}_{e} = \lVert(\beta\cdot \mathbf{E} + \mathbf{1})\odot (\mathbf{T}_{adv} - \mathbf{T}_{0})\rVert^2_2,
    \end{split}
    \label{eqn:aeloss}
\end{equation}
where the $\beta\cdot \mathbf{E} + \mathbf{1}$ is the weight tensor, the $\mathbf{1}$ is a tensor in which each element is 1 and its dimension is same with $\mathbf{E}$ and $\odot$ denotes the element-wise multiplication.

To further improve the naturalness of the camouflage, we introduce the smooth loss \cite{Eykholt_2018_CVPR} by reducing the difference square between adjacent pixels. For a rendered adversarial image $\mathbf{I}_{adv}$, the smooth loss can be formulated as:

\begin{equation}
    \begin{split}
        \mathcal{L}_{s} = \sum (x_{i,j}-x_{i+1,j})^2 + (x_{i,j}-x_{i,j+1})^2,
    \end{split}
    \label{eqn:smoothloss}
\end{equation}
where $x_{i,j}$ is the pixel value of $\mathbf{I}_{adv}$ at coordinate $(i,j)$.

To sum up, the generated camouflage in this case will be visually correlated to the scenario context in both the pixel and perceptual level, leading to evade the human perceptual attention. 

\subsection{Overall Optimization Process}
Overall, we generate the adversarial camouflage by jointly optimizing the model attention distraction loss $\mathcal{L}_{d}$, human attention evasion loss $\mathcal{L}_{e}$, and smooth loss $\mathcal{L}_{s}$. 

Specifically, we first distract the target model from the salient objects to the meaningless part (\eg, background); we then evade the human-specific attention mechanism by enhancing the strong perceptual correlation to the scenario context. Thus, we can generate transferable and visually-natural adversarial camouflages by minimizing the following formulation as
\begin{equation}
\label{equ:total-loss}
    \begin{split}
    \min\mathcal{L}_{d} + \lambda\mathcal{L}_{e} +\mathcal{L}_{s},
    \end{split}
\end{equation}
where $\lambda$ controls the contribution of the term $\mathcal{L}_{e}$.

To balance the attacking ability and appearance naturalness, we set $\lambda$ as $10^{-5}$ in the classification task and $10^{-3}$ in the detection task, and set $\beta$ as 8 according to our experimental results. The overall training algorithm can be described as Algorithm \ref{alg:alg1}.

\begin{algorithm}[htb]
\caption{Dual Attention Suppression (DAS) Attack}
\label{alg:alg1}
\begin{algorithmic}
\renewcommand{\algorithmicrequire}{\textbf{Input:}}
\renewcommand{\algorithmicensure}{\textbf{Output:}}
\REQUIRE environmental parameter set $C=\{c_1,c_2,...c_{r}\}$ , 3D real object $(\mathbf{M}, \mathbf{T})$, seed content patch $\mathbf{P}_{0}$, neural renderer $\mathcal{R}$, attention model $\mathcal{A}$, and a class label $y$
\ENSURE adversarial texture tensor $\mathbf{T}_{adv}$
\STATE $\mathbf{T}_{0} \gets \Psi(\mathbf{P}_{0}, \mathbf{T})$
\STATE $\mathbf{P}_{edge} \gets \Phi(\mathbf{P}_{0})$
\STATE transform $\mathbf{P}_{edge}$ to $\mathbf{E}$
\STATE initial $\mathbf{T}_{adv}$ as $\mathbf{T}_{0}$
\FOR{the number of epochs}
\STATE select $minibatch$ environmental conditions from $C$
\FOR{$m = r/minibatch$ steps}
\STATE $\mathbf{I}_{adv} \gets \mathcal{R}((\mathbf{M},\mathbf{T}_{adv}), c_{m})$
\STATE $\mathbf{S}^{y} \gets \mathcal{A}(\mathbf{I}_{adv}, y)$
\STATE calculate the $\mathcal{L}_{d}$, $\mathcal{L}_{e}$ and $\mathcal{L}_{s}$ by Eqn (\ref{eqn:adloss}, \ref{eqn:aeloss}, \ref{eqn:smoothloss})
\STATE optimize the $\mathbf{T}_{adv}$ by Eqn (\ref{equ:total-loss})
\ENDFOR
\ENDFOR
\end{algorithmic}
\end{algorithm}

\section{Experiments}
In this section, we first outline the experimental settings, we then illustrate the effectiveness of our proposed attacking framework by thorough evaluations in both the digital and physical world.

\subsection{Experimental Settings}
\textbf{Virtual environment}.
To perform a physical world attack, we choose CARLA \cite{Dosovitskiy17} as our 3D virtual simulated environment, which is the commonly used open-source simulator for autonomous driving research. Based on Unreal Engine 4, CARLA provides many high-resolution open digital assets, \eg, urban layouts, buildings, and vehicles to simulate a digital world that is nearly the same as the real world. 

\textbf{Evaluation metrics}.
To evaluate the performance of our proposed method, we select the widely used $\mathrm{Accuracy}$ as the metric for the classification task; as for the detection task, we adopt the P@0.5 following \cite{zhang2018camou}, which reflects both the IoU and precision information. 

\textbf{Compared methods}. We choose several state-of-the-art works in the 3D attack and physical attack literature, including UPC \cite{Huang_2020_CVPR}, CAMOU \cite{zhang2018camou}, and MeshAdv \cite{Xiao_2019_CVPR}. For better analysis, we implement MeshAdv in different ways. We use ResNet-50 as its base-model for the classification and Yolo-V4 for detection. We provide more information about these methods in Supplementary Material.

\textbf{Target models}. We select different commonly used model architectures for experiments. Specifically, Inception-V3 \cite{articleIV3}, VGG-19 \cite{vgg19}, ResNet-152 \cite{ResNet50}, and DenseNet \cite{DenseNet} are employed for the classification task; Yolo-V5 \cite{yolo}, SSD \cite{ssd}, Faster R-CNN \cite{faster-rcnn}, and Mask R-CNN \cite{mask-rcnn} are employed for the detection task. For all the models, we use the pre-trained version on ImageNet and COCO. 

\textbf{Implementation details}. We empirically set $\lambda = 10^{-5}$ for classification task, $\lambda = 5\times10^{-3}$ for detection task and we set $\beta = 8$. We adopt an Adam optimizer with a learning rate of 0.01, a weight decay of ${10}^{-4}$, and a maximum of 5 epochs. We employ a seed content patch (\eg, a stick smile face image) as the appearance of the 3D object in the training process. All of our codes are implemented in PyTorch. We conduct the training and testing processes on an NVIDIA Tesla V100-SXM2-16GB GPU cluster. In the physical world attack scenario, adversaries only have limited knowledge and access to the deployed models (\ie, architectures, weights, \etc.). Considering this, we mainly focuses on attacks in the black-box settings, which is more meaningful and applicable for physical world attacks.

\subsection{Digital World Attack}
In this section, we evaluate the performance of our generated adversarial camouflages on the vehicle classification and detection task in the digital world under black-box settings.

We randomly select 155 points in the simulation environment to place the vehicle and use a virtual camera to capture 100 images at each point using different settings (\ie, angles, and distances). Specifically, we use different distance values (5, 10, 15, and 20), four camera pitch angle values (22.5$^{\circ}$, 45$^{\circ}$, 67.5$^{\circ}$, and 90$^{\circ}$), and eight camera yaw angle values (south, north, east, west and southeast, southwest, northeast, northwest). 
We then collect 15,500 simulation images with different setting combinations, and we choose 12,500 images as the training set and 3,000 images as the test set.
To conduct fair comparisons, we use the backbone of ResNet-50 (for classification) and Yolo-V4 (for detection) as attention modules in training. As illustrated in Table \ref{tab:classification} and Table \ref{tab:detection}, we can draw several conclusions as follows:

(1) Our adversarial camouflage achieves significantly better performance for both classification and detection tasks on different models (a maximum drop by \textbf{41.02\%} on ResNet-152 and a maximum drop by \textbf{23.93\%} on Faster R-CNN).

(2) We found that UPC works comparatively worse than other baselines for detection task. We conjecture the reason might be that UPC is primarily designed for physical attacks therefore showing worse attacking ability in the digital world. By contrast, our DAS attack exploits the intrinsic characteristics, which still achieves good attacking ability in the digital world.

(3) SSD shows evidently better robustness compared to other backbone models (\ie, lower accuracy decline). The reason might be that some modules in SSD are less vulnerable to adversarial attacks, which could be used to further improve model robustness. We put it as future work. 

\begin{table}[!tbh]
\footnotesize
\begin{center}

\begin{tabular}{ccccc}
\hline
\multirow{2}{*}{\centering\textbf{Method}}&\multicolumn{4}{c}{Accuracy (\%)}\\
\cline{2-5}
& Inception-V3&VGG-19&ResNet-152&DenseNet\\
\hline
Raw& 74.36&40.62&73.51&71.91\\
\hline
MeshAdv&42.31&32.44&35.33&58.04\\
\hline
CAMOU&47.51&31.46&48.93&57.56\\
\hline
 UPC&42.40&38.00&48.18&65.87\\
\hline
 Ours&\textbf{39.86}&\textbf{30.18}&\textbf{32.49}&\textbf{55.42}\\
\hline
\end{tabular}
\end{center}
\caption{The results in the digital world on the classification task.}
\label{tab:classification}
\end{table}

\begin{table}[!tbh]
\footnotesize
\begin{center}
\begin{tabular}{ccccc}
\hline
{\multirow{2}{*}{\centering\textbf{Method}}}&\multicolumn{4}{c}{P@0.5 (\%)}\\
\cline{2-5}
& Yolo-V5&SSD&Faster R-CNN&Mask R-CNN\\
\hline
Raw&92.07&81.54&86.04&89.24\\
\hline
MeshAdv&72.45&66.44&71.84&80.84\\
\hline
CAMOU&74.01&73.81&69.64&76.44\\
\hline
UPC&82.41&74.58&76.94&81.97\\
\hline
Ours&\textbf{72.58}&\textbf{65.81}&\textbf{62.11}&\textbf{70.21}\\
\hline
\end{tabular}
\end{center}
\caption{The results in the digital world on the detection task.}
\label{tab:detection}
\end{table}
\subsection{Physical World Attack}

\begin{figure}[tbh]
\begin{center}
\includegraphics[width=0.9\linewidth]{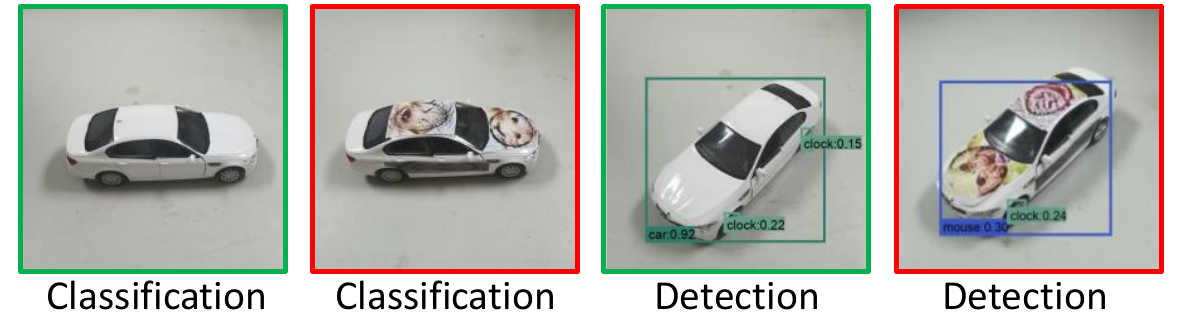}
\end{center} 
\caption{The results of attacking toy cars. They are respectively recognized as \texttt{car}, \texttt{sandal}, \texttt{car}, \texttt{mouse}.}
\label{fig:phy-ours}
\end{figure}

As for the physical world attack, we conduct several experiments to validate the practical effectiveness of our generated adversarial camouflages. Due to the limitation of funds and conditions, we print our adversarial camouflages by an HP Color LaserJet Pro MFP M281fdw printer and stick them on a toy car model with different backgrounds to simulate the real vehicle painting. To conduct fair comparisons, we take 144 pictures of the car model on various environmental conditions (\ie, 8 directions \{left, right, front, back and their corresponding intersection directions\}, 3 angles \{0$^{\circ}$, 45$^{\circ}$, 90$^{\circ}$\}, 2 distances \{long and short distances\} and 3 different surroundings) using a Huawei P40 phone. The visualization of our generated adversarial camouflages can be found in Figure \ref{fig:phy-ours}. 

The evaluation results can be witnessed in Table \ref{tab:phy-cls} and Table \ref{tab:phy-det}. Compared with other methods, the DAS shows competitive transferable attacking ability, which is significantly better than the compared baselines (\eg, \textbf{31.94\%} on Inception-V3, \textbf{27.78\%} on VGG-19, \textbf{29.86\%} on ResNet-152, and \textbf{34.03\%} on DenseNet,  respectively). Moreover, the evaluation result of UPC appears a distinct improvement than that in the digital world, which is consistent with our analysis. However, the SSD shows lower robustness in the physical world which is worth further study. Besides, the Yolo-V5 shows stunning P@0.5 values, which probably because that Yolo-V5 is specially designed for applications in the physical world. Though facing this strong model, our DAS method still shows a certain attacking ability compared with others.

To sum up, the experimental results demonstrate the strong transferable attacking ability of our adversarial camouflages in the physical world.

\begin{table}[b]
\footnotesize
\begin{center}

\begin{tabular}{ccccc}
\hline
\multirow{2}{*}{\centering\textbf{Method}}&\multicolumn{4}{c}{Accuracy (\%)}\\
\cline{2-5}
& Inception-V3&VGG-19&ResNet-152&DenseNet\\
\hline
Raw& 58.33&40.28&41.67&46.53\\
\hline
MeshAdv&40.28&34.03&38.89&36.11\\
\hline
CAMOU&40.28&29.17&31.25&45.14\\
\hline
 UPC&35.41&33.33&33.33&41.67\\
\hline
 Ours&\textbf{31.94}&\textbf{27.78}&\textbf{29.86}&\textbf{34.03}\\
\hline
\end{tabular}
\end{center}
\caption{The results in the physical world on the classification task.}
\label{tab:phy-cls}
\end{table}

\begin{table}[!tbh]
\footnotesize
\begin{center}
\begin{tabular}{ccccc}
\hline
{\multirow{2}{*}{\centering\textbf{Method}}}&\multicolumn{4}{c}{P@0.5 (\%)}\\
\cline{2-5}
& Yolo-V5&SSD&Faster R-CNN&Mask R-CNN\\
\hline
Raw&100.00&90.28&68.06&93.75\\
\hline
MeshAdv&100.00&61.11&56.25&63.19\\
\hline
CAMOU&99.31&61.11&61.81&63.19\\
\hline
UPC&100.00&63.19&52.08&61.81\\
\hline
Ours&\textbf{92.36}&\textbf{56.25}&\textbf{44.44}&\textbf{54.86}\\
\hline
\end{tabular}
\end{center}
\caption{The results in the physical world on the detection task.}
\label{tab:phy-det}
\end{table}

\subsection{Model Attention Analysis}

In this part, we conduct a detailed analysis on model attention through both qualitative and quantitative studies to validate the effectiveness the model attention distraction in our DAS attack. 

\begin{figure}[b]
\subfigure[]{
\includegraphics[width=0.40\linewidth]{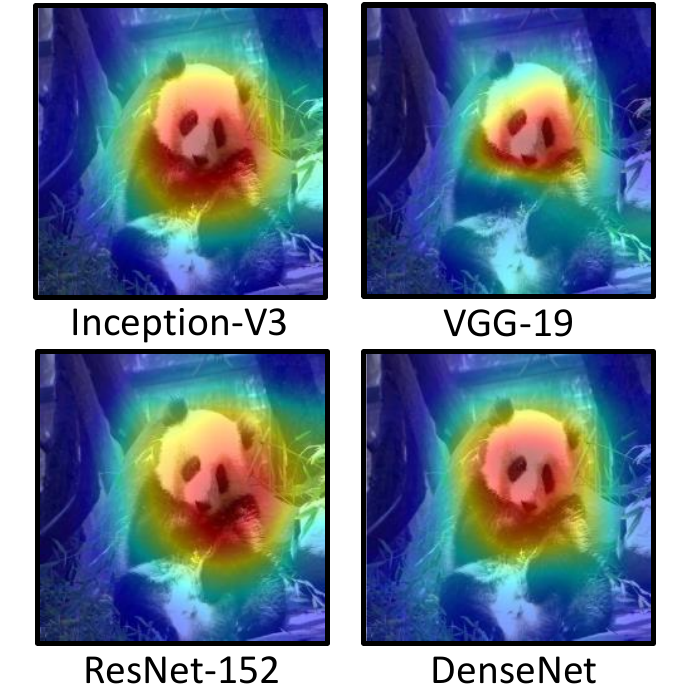}
 \label{fig:diff-att}
}
\subfigure[]{
\includegraphics[width=0.50\linewidth]{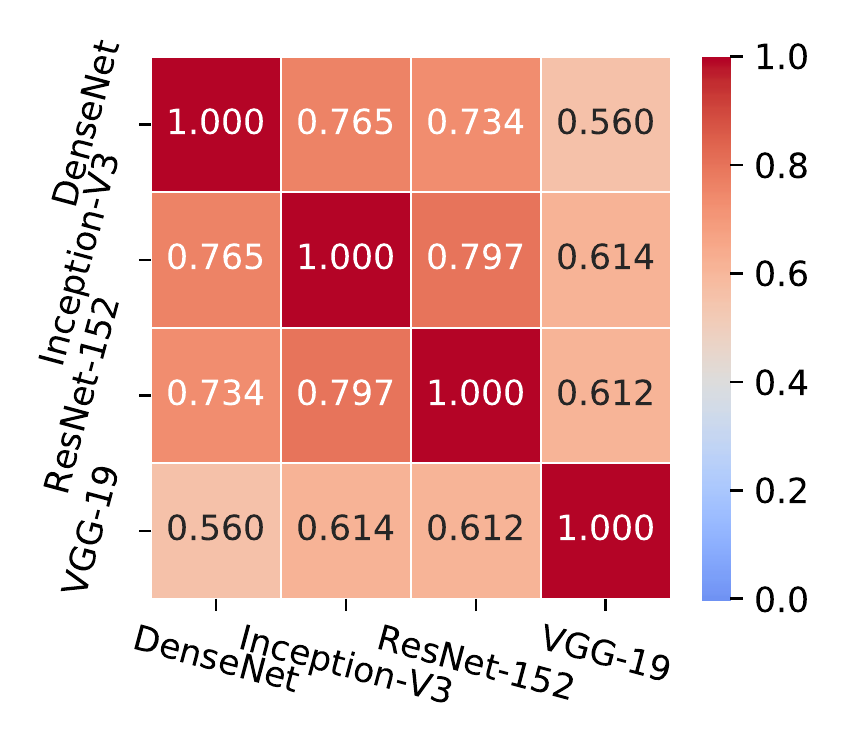}
 \label{fig:ssim}
}
\caption{(a) is the attention maps on 4 different models to a particular image. (b) is a heat map drawn according to the SSIM values.}
\end{figure}

Firstly, we conduct a qualitative study by visualizing the attention regions of different models towards the same image. As shown in Figure \ref{fig:diff-att}, different DNNs show similar attention patterns towards the same image. In other words, different models pay their attention to similar regions, indicating that the attention is shared among models and can be deemed as a model-agnostic characteristic. 

We then conduct a quantitative study by calculating the structural similarity index measure (SSIM) \cite{ssim}, which is a well-known quality metric used to measure the similarity between two images \cite{ssimeva}. Specifically, we generate the attention maps of a specific image (\ie, panda) on different models and calculate the SSIM values between each pair of the attention maps on different models (\ie, Inception-V3, VGG-19, ResNet-152, and DenseNet). As shown in Figure \ref{fig:ssim}, different models demonstrate comparatively high similarities of the attention maps.

\begin{figure}[t]
\begin{center}
\includegraphics[width=0.9\linewidth]{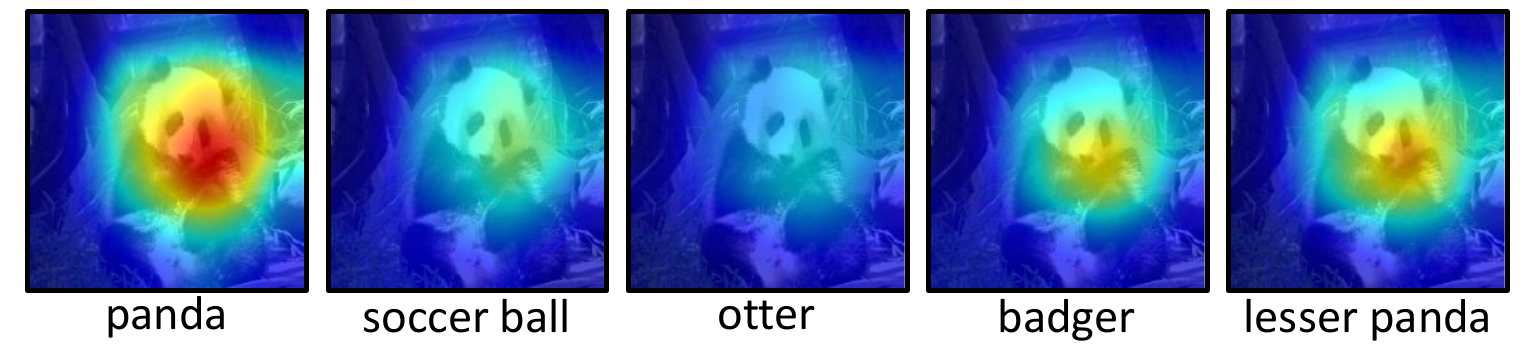}
\end{center}
   \caption{The visualization of attention maps on the same image \texttt{panda} with different target labels using ResNet-152. The attention maps differ significantly when different target labels are provided to the model.}
\label{fig:diff-view}
\end{figure}

In addition, we further visualize the attention maps by changing the model predictions (\ie, class). As shown in Figure \ref{fig:diff-view}, when changing the class label, the attention map is distracted from the salient objects and becomes more sparse over the entire image.

\begin{figure}[!htb]
\begin{center}
\includegraphics[width=0.9\linewidth]{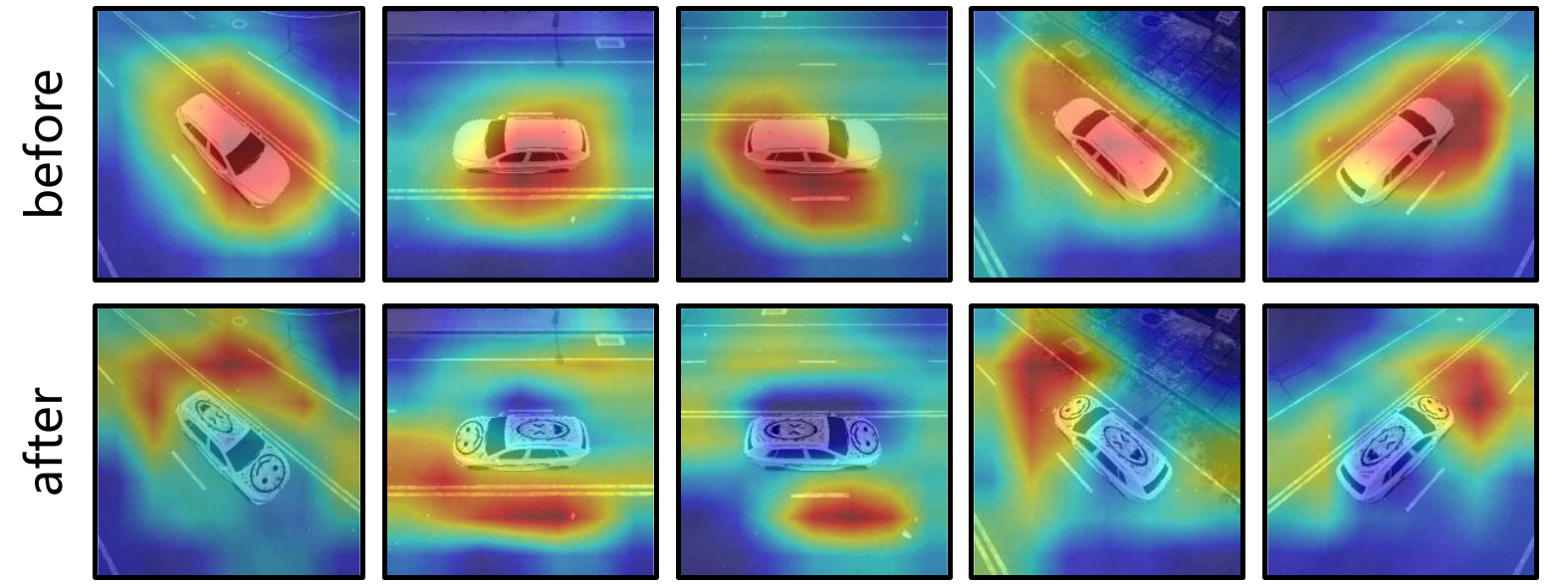}
\end{center} 
   \caption{The attention maps before and after our DAS attack. After our DAS attack, the model attention is distracted.}
\label{fig:att-results}
\end{figure}

In summary, we can draw several conclusions as follows: (1) different DNNs show similar attention patterns towards the same class in a specified image; (2) we can adversarially attack a DNN to wrong predictions by distracting its attention. More experimental results can be found in the Supplementary Material.

\subsection{Human Perception Study}

To evaluate the naturalness of our generated adversarial camouflage, we conduct a human perception study on one of the most commonly used crowdsourcing platform. We adversarially perturb our 3D car object using different methods (\ie, MeshAdv, CAMOU, UPC, and Ours) and get the adversarial textures. Then we paint the car using these camouflages and get the rendered images for human perception studies as follows: (1) Recognition. The participants are asked to assign each of the camouflages generated by the methods above to one of the 8 classes (the ground-truth class, 6 classes similar to the ground-truth, and ``I cannot tell what it is''). As for CAMOU, given it lacks semantic information, we do not consider it for the recognition task; (2) Naturalness. The participants are asked to score the naturalness of the camouflages from 1 to 10. In particular, we collect all responses from 106 participants. 

\begin{table}[tbh]
\footnotesize
\begin{center}
\begin{tabular}{ccccc}
\hline
\multirow{2}{*}{\centering\textbf{Question}}&\multicolumn{4}{c}{Percent (\%)}\\
\cline{2-5}
& MeshAdv&CAMOU&UPC&Ours\\
\hline
Recognition&36.6&--&27.4&\textbf{49.6}\\
\hline
Naturalness&43.4&39.6&40.6&\textbf{60.4}\\
\hline
\end{tabular}
\end{center}
\caption{The results of human perception study.}
\label{tab:human}
\end{table}

As shown in Table \ref{tab:human}, \textbf{49.6\%} of the participants can recognize the ground-truth label for our camouflages, which are far better than those generated by other methods. As for the naturalness task, up to \textbf{60.4\%} of the participants believe that our adversarial camouflage is natural-looking, which outperforms others by large margins (17\%+). Thus, we can conclude that our adversarial camouflage is most visually natural and perceptually consistent to human perception. \footnote{Our experimental details can be accessed at \url{https://github.com/nlsde-safety-team/DualAttentionAttack}.}

\subsection{Ablation Studies}
In this section, we conduct several ablation studies to further investigate the contributions of our two main loss terms, \ie, the model attention distraction loss and the human attention evasion loss. Due to the fact that the smooth loss is fully studied in \cite{Eykholt_2018_CVPR}, we set it as a fixed term.

\textbf{The effect of different loss terms}. 
Different loss terms play different roles, we conduct an ablation study to further investigate the effect of loss terms. We argue that the model attention distraction loss $\mathcal{L}_{d}$ mainly provides a transferable attacking ability in our DAS method and the human attention evasion provides the natural appearance. To prove these views, we conduct an experiment by calculating different loss term combinations. Specifically, we optimize the adversarial camouflage using function $\mathcal{L}_{d}$, $\mathcal{L}_{e}$, and $\mathcal{L}_{d}+\lambda\mathcal{L}_{e}$ respectively (with $\mathcal{L}_{s}$ fixed). 
As shown in Table \ref{tab:loss}, the accuracy shows a significant drop (\ie, \textbf{36.53\%} under $\mathcal{L}_{d}$ setting to 59.87\% under $\mathcal{L}_{e}$ setting, 39.86\% under $\mathcal{L}_{d}+\mathcal{L}_{e}$ setting). And the corresponding SSIM values generated with a benign image are 0.6905, 0.9987, and 0.7551 respectively, demonstrating our viewpoints. Besides, an interesting result can be observed in our experiments. When training under $\mathcal{L}_{e}$ setting, the accuracy appears an evident improvement on VGG-19 and DenseNet but drop on Inception-V3 and ResNet-152, which means that common textures may cause agnostic impact to DNNs, further demonstrating their vulnerability.

\begin{table}[!tbh]
\footnotesize
\begin{center}
\begin{tabular}{ccccc}
\hline
\multirow{2}{*}{\centering\textbf{Method}}&\multicolumn{4}{c}{Accuracy (\%)}\\
\cline{2-5}
& Inception-V3&VGG-19&ResNet-152&DenseNet\\
\hline
Raw & 74.36&40.62&73.51&71.91\\
\hline
$\mathcal{L}_{d}$& 36.53&25.87&31.20&51.73\\
\hline
$\mathcal{L}_{e}$&59.87&50.00&47.87&75.07\\
\hline
$\mathcal{L}_{d}+\lambda\mathcal{L}_{e}$&39.86&30.18&32.49&55.42\\
\hline
\end{tabular}
\end{center}
\caption{The ablation study on attention distraction portion. We set $\lambda$ as $10^{-5}$.}
\label{tab:loss}
\end{table}

\begin{figure}[t]
\begin{center}
\includegraphics[width=0.8\linewidth]{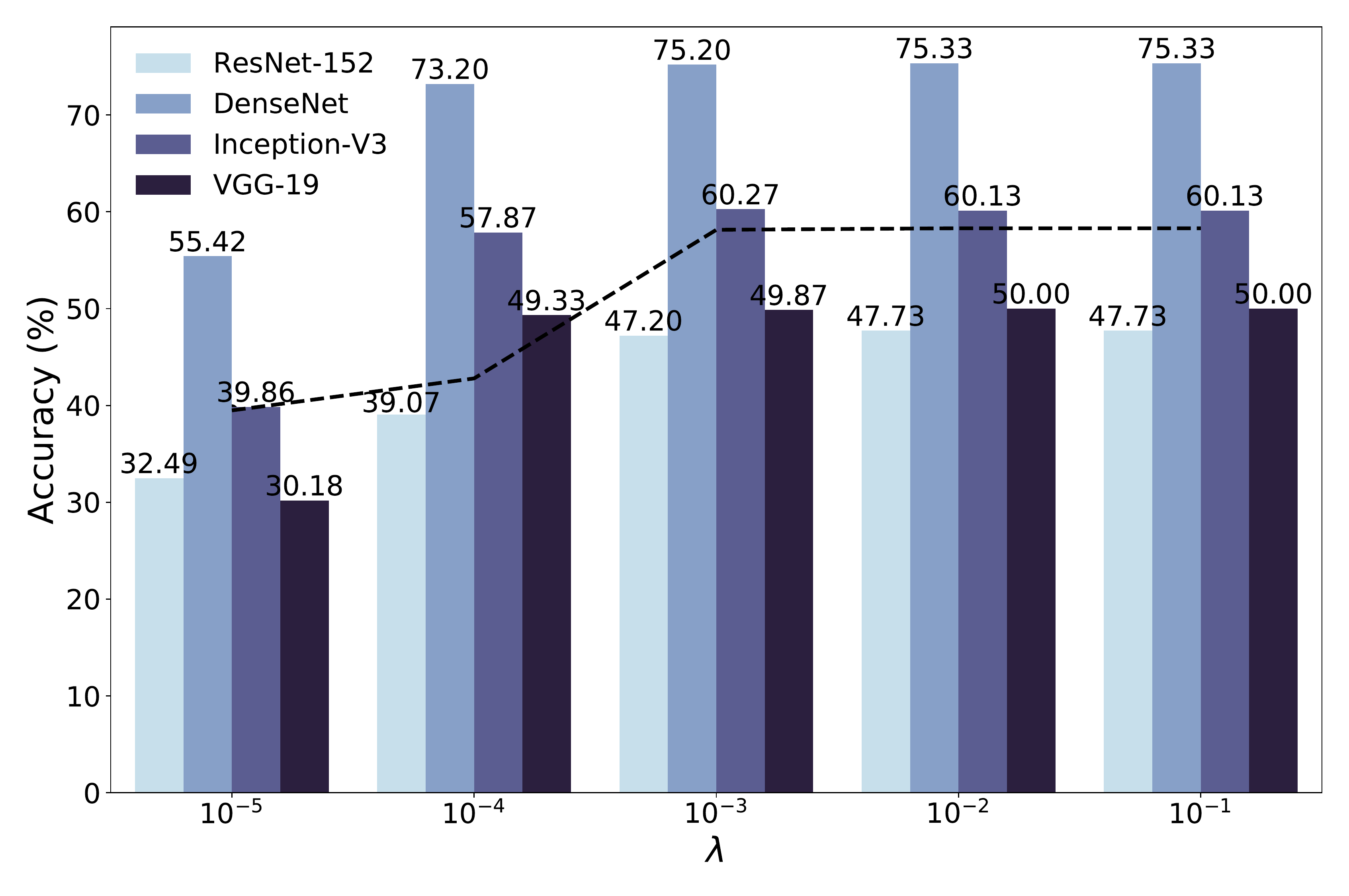}
\end{center} 
\caption{Ablation on studying the effectiveness of $\lambda$. The dotted line represents the trend of accuracy change, and the corresponding value of each $\lambda$ is the average accuracy of the four models.}
\label{fig:ablation-lambda}
\end{figure}

\textbf{The effect of hyper-parameter $\lambda$}. 
Regarding the hyper-parameter $\lambda$, we argue that it controls the level of the strong semantic correlation with the scenario context.
We evaluate the effectiveness of $\lambda$ on a ResNet-50 model using Accuracy and SSIM. Specifically, we set the $\lambda$ as $10^{-5}$, $10^{-4}$, $10^{-3}$, $10^{-2}$, and $10^{-1}$, respectively. As illustrated in Figure \ref{fig:ablation-lambda}, the model accuracy first increases and then keeps a stable value as $\lambda$ increases. We calculate the SSIM values between each pair of the clean and corresponding adversarial example, which shows the similar tendency (\ie, 0.7034, 0.7551, 0.8750, 0.9982, 0.9991, and 0.9998, the closer to 1 the SSIM value is, the more similar the images are). According to the results, we can draw the conclusion that $\lambda$ balances the attacking ability and appearance. When $\lambda$ gets bigger, the accuracy and SSIM value get bigger, which means lower attacking ability and better appearance. And finally, the SSIM achieves its upper bound, leading to the loss of additional attacking ability.

\section{Conclusion}

In this paper, we propose the Dual Attention Suppression (DAS) attack to generate adversarial camouflage in the physical world by suppressing both model and human attention. To improve the transferability of adversarial camouflages, we suppress the model attention by distracting the model-shared similar attention from target to non-target regions. Since our generated camouflage captures the model-agnostic structures, it can transfer among different models. To generate more visually-natural camouflage, we suppress the human attention by evading the human-specific bottom-up attention. By preserving the shape of a seed content patch which has strong semantic association to the scenario context, the generated camouflage can be highly correlated to human perception, which is more natural and unsuspicious to human attention. We conduct extensive experiments for both classification and detection tasks in both the digital and physical world under black-box setting, and our DAS outperforms state-of-the-art baselines.

In the future, we are interested in investigating the attack abilities of our adversarial camouflage using a real vehicle in the real-world scenario. Using projection or 3D printing, we could simply paint our camouflage on a real-world vehicle. Further, we would also like to investigate the effectiveness of our generated camouflage to improving model robustness against different noises.

{\small
\bibliographystyle{ieee_fullname}
\bibliography{egbib}
}

\end{document}